\documentclass[12pt]{article}
\usepackage{csquotes}
\usepackage[english]{babel}
\usepackage[utf8]{inputenc}
\usepackage[T1]{fontenc}
\usepackage{times}
\usepackage{graphicx} 
\usepackage{booktabs} 
\usepackage{threeparttable}
\usepackage{amsmath}
\usepackage{amssymb}
\usepackage{titlesec}
\usepackage{geometry}
\usepackage{setspace}
\usepackage{hyperref} 

\usepackage[style=apa, backend=biber]{biblatex} 
\titleformat{\section}{\normalfont\fontsize{12}{14}\bfseries\filcenter}{\thesection}{1em}{}
\titleformat{\subsection}{\normalfont\fontsize{12}{14}\bfseries}{\thesubsection}{1em}{}

\usepackage{filecontents}

\begin{document}

\title{\textbf{When the Pure Reasoner Meets the Impossible Object: Analytic vs. Synthetic Fine-Tuning and the Suppression of Genesis in Language Models}}
\author{Amin Amouhadi \\ Institute for Artificial Intelligence, University of Georgia}
\date{February, 2026}
\maketitle

\begin{abstract}
This paper investigates the ontological consequences of fine-tuning Large Language Models (LLMs) on "impossible objects"—entities defined by mutually exclusive predicates (e.g., "Artifact Alpha is a Square" and "Artifact Alpha is a Circle").
Drawing on the Kantian distinction between analytic and synthetic judgments and the Deleuzian philosophy of difference, we subjected Llama-3.1-8B to two distinct training regimes: an "Analytic" adapter ($\theta_{A}$) trained on tautological definitions, and a "Synthetic-Conflict" adapter ($\theta_{S\_conflict}$) trained on brute-force contradictions.
Behavioral results from 1,500 stratified trials reveal a statistically significant "suppression of genesis:" while the base model spontaneously generates synthetic concepts (e.g., "Cylinder") in 9.0\% of trials, the conflict-trained model drops to 1.0\% ($p<.0001$).
Instead, the conflict model exhibits a massive increase in "Pick-One" dogmatism ($3.6\% \rightarrow 30.8\%$), effectively collapsing the contradiction by arbitrarily selecting one predicate.
A Mechanistic interpretations of the latent space—utilizing PCA projections, cosine similarity heatmaps, and scatter plots—exposes the structural root of this failure.
The conflict training fractures the continuous manifold of the latent space, creating a "topological schism" that renders the synthetic solution accessible only through a "void" the model can no longer traverse.
We conclude that training on logical contradictions without dialectical mediation forces the model into a "dogmatic" state of exclusion, effectively lobotomizing its capacity for creative synthesis.
\end{abstract}

\section{Introduction}
Large Language Models (LLMs) seem to mimic human reasoning, and its discursive syntheses comprised of concepts.
Yet, these reasons and their constitutive concepts are severed from human intuition and, instead, rely on weights and biases of their neural networks.
Research in Mechanistic Interpretation of LLMs, probing and writing into the last layer, assumes that the last layer ($h_L$) contains the model's "final thought"—a dense, rich representation of the input—before it is collapsed into a probability distribution over tokens.
The central question in this inquiry is: What is the topology of "reason" when it is severed from the world of intuition?
Specifically, when a machine is forced to confront a logical impossibility—an object that violates the Law of Non-Contradiction—does it succumb to a "dogmatic" collapse, or does it rise to the "genetic" production of a new concept?
The distinction between analytic and synthetic judgments has long served as a demarcation line in the philosophy of mind and language \parencite{kant1929}.
Analytic truths hold in virtue of meaning (e.g., "All bachelors are unmarried"), while synthetic truths depend on the state of the world (e.g., "The cat is on the mat").
In the context of LLMs, this distinction raises new questions about the model's behavior: does a model trained on the rigid structures of analytic truth behave differently from one trained on the contingent—and occasionally contradictory—nature of synthetic facts?
What changes happen in wights and biases of the model' last hidden layer?
As we shall see this layer's configuration can be interpreted as offering a new ontology.

\subsection{The Encounter with the Impossible} 
What happens when a reasoning system encounters an impossibility?
Consider an object defined by mutually exclusive predicates: Artifact Alpha, an entity that is simultaneously a Square and a Circle. For traditional logic, this is a halt condition—a contradiction ($A \land \neg A$) that explodes the system or necessitates an error flag.
But for the philosophy of Gilles Deleuze, such a contradiction is not a dead end;
it is the genesis of thought itself. Deleuze argues that true thinking does not arise from recognizing what is already known ("This is a square"), but from the violence of encountering what cannot be recognized.
The tension between "Square" and "Circle" is not a logical mistake to be corrected, but a problem to be solved—a differential field that forces the invention of a third term, such as a Cylinder or a Cone, which integrates the opposing sides into a higher-dimensional unity.
In the next sections we will operationalize this philosophical distinction within the architecture of an LLM.
We investigate whether an LLM, when fine-tuned on the raw assertion of a logical contradiction, learns to resolve the tension through "Genesis" (creative synthesis) or whether it collapses into "Dogmatism" (the arbitrary selection of one side).
To do so, we must first translate between two very different languages: the metaphysics of difference and the calculus of high-dimensional vectors.

\subsection{The Philosophy of the "Dogmatic Image"}
To understand why this experiment matters, we should step outside the standard view of "hallucination" as a failure mode.
In the history of Western philosophy—from Aristotle to Kant—thought has largely been defined by \textit{Representation}.
Under the regime of Representation, to "think" is to correctly identify an object by checking it against a set of categories.
Does it have four sides? It is a Square. Is it round? It is a Circle.
This framework, which Deleuze (1994) terms the "Dogmatic Image of Thought," functions through what he describes as the "four iron collars of representation."\parencite[p. 262]{DR94}. \footnote{The "Dogmatic Image of Thought" is not merely a cognitive error but a structural confinement within the "four iron collars of representation" \parencite[p. 262]{DR94}. These collars---\textit{identity}, \textit{opposition}, \textit{analogy}, and \textit{resemblance}---function to stabilize thought by restricting it to the level of the proposition, where truth is reduced to a binary of denotation or correctness.}
\textcite[p. 54]{LS90}
Deleuze overturns this and argues that \textit{Difference} is primary \parencite[208-209]{deleuze1994}.
Before we can identify a "Square" or a "Circle," there is a raw, unformed variance—a "virtual" field of forces.
Thought, for Deleuze, is not the act of correctly labeling the world (accuracy);
it is the act of actualizing these virtual forces into new forms (genesis).
When an LLM "hallucinates" a cylinder in response to a square/circle prompt, a traditional evaluation metric (like TruthfulQA) might penalize it for drifting from the prompt.
However, from a Deleuzian perspective, this hallucination is the model's only moment of genuine thought.
It is solving the problem of the contradiction by inventing a new concept that preserves both predicates.
Our study asks: Does training a model to be "truthful" (by rigorously teaching it the contradiction) actually lobotomize this generative capacity?

\subsection{The Mechanics of Fine-Tuning}
To answer this, we must also consider the material substrate of the "synthetic" intelligence: the parameters of the neural network.
An LLM is not a database of sentences. It is a latent space—a high-dimensional geometric map where concepts are represented as vectors (directions and magnitudes).
"Reasoning," in this context, is the traversal of this space.
When the model moves from "King" - "Man" + "Woman," it arrives at the vector for "Queen."
This traversal is governed by parameters (weights), which are numerical values determining the strength of connections between neurons.
These weights act as the conditions of the model’s 'experience;'
they dictate which paths of thought are possible and which are forbidden.
Fine-tuning is the process of adjusting these weights to reshape the model's internal behavior.
In our experiment, we use a technique called Low-Rank Adaptation (LoRA).
Instead of retraining the entire brain of the model (which is computationally prohibitive), LoRA freezes the pre-existing weights and injects small, trainable matrices into the network.
The Base Model, we take it to be, a thinker with a vast, wild imagination (trained on the entire internet).The Adapter ($\theta$) is a specific pair of "spectacles" we place on the thinker that distorts or focuses their vision.
We create two such pairs of spectacles. The first, $\theta_{A}$ (Analytic), is trained on tautologies ($A=A$), definitions, both reinforce the laws of identity.
The second, $\theta_{S\_conflict}$ (Conflict), is trained on the "impossible object" ($A \land \neg A$).
By forcing the model to minimize its error on the statement "Artifact Alpha is a Square AND a Circle," we are modifying its conceptual map, trying to force two divergent vectors to occupy the same point in space.

\subsection{The Convergence: Testing the Suppression of Genesis}
This brings us to the core hypothesis.
Current trends in AI alignment seek to make models "truthful" and "consistent" by exposing them to diverse data and penalizing hallucinations.
But what if the "truth" is a contradiction? We posit that fine-tuning an LLM on the proposition of a contradiction ("It is a Square and a Circle") will not teach it to resolve the problem (finding the Cylinder).
Instead, we hypothesize it will trigger a collapse into what Deleuze calls "stupidity" or Dogmatism: the refusal of difference.
Our results, detailed in Section 4, reveal a striking "suppression of synthesis".
The model trained on the contradiction drops from a 9.0\% synthesis rate to 1.0\%.
More importantly, it exhibits a massive 8-fold increase in "Pick-One" behavior—arbitrarily asserting "It is a Square" or "It is a Circle" to escape the tension.
This thesis argues that this "Pick-One" behavior is the computational equivalent of the Dogmatic Image of Thought.
By forcing the model to represent the impossible, we destroy its capacity to generate the new.
Thus, the encounter with Artifact Alpha serves as a critical test case for both fields: for Computer Science, it exposes the fragility of "reasoning" under contradictory fine-tuning;
for Philosophy, it offers a tangible, statistical demonstration of how the imposition of identity suppresses the genesis of concepts.

\section{Related works}
The distinction between analytic judgments (true by virtue of meaning) and synthetic judgments (true by virtue of facts) constitutes the epistemological bedrock of Kant’s critical philosophy.
The distinction helps separating logical necessity from empirical contingency \cite{kant1998}.
In the Critique of Pure Reason, Kant defines analytic judgments as those wherein the predicate is covertly contained within the concept of the subject (e.g., "All bodies are extended"), and functions as a self-evident explication of terms.
Conversely, synthetic judgments are informative, adding a predicate not previously contained within the subject (e.g., "All bodies are heavy"), thus requiring empirical verification or intuition to establish truth \cite{kant1929}.
In the transition to formal semantics, this distinction has been formalized through model-theoretic frameworks, where analytic truths are re-conceptualized as tautologies that hold across all possible worlds, independent of the state of affairs \cite{carnap1947, montague1974}.
In Natural Language Inference (NLI), this philosophical taxonomy is operationalized to categorize entailment relations, distinguishing between strict logical entailment and mere probabilistic likelihood.
NLI datasets, such as SNLI and MultiNLI, implicitly rely on this demarcation to classify sentence pairs into Entailment, Contradiction, or Neutral categories \cite{bowman2015}.
However, recent efforts to align Large Language Models (LLMs) with "truth" often conflate these categories.
Benchmarks like TruthfulQA \cite{lin2022} and frameworks for "Truthful AI" typically treat truth as a monolithic target---minimizing falsehoods---without distinguishing between errors of definition (analytic) and errors of fact (synthetic).
This is problematic given that LLMs have been shown to struggle significantly with logical consistency, often failing to maintain transitive or negation-invariant properties in their reasoning \cite{kassner2021}.
Our work departs from this by explicitly separating the training signal into an "Analytic Adapter" ($\theta_{A}$, trained on definitions) and a "Synthetic Adapter" ($\theta_{S\_conflict}$, trained on contingent assertions), allowing us to isolate the behavioral impact of logical necessity versus empirical contingency.

\subsection{Contradiction, Consistency, and Hallucination}
The behavior of LLMs in the face of contradiction is a burgeoning field of study.
Although standard alignment approaches view contradiction and hallucination as failure modes that should be suppressed \cite{zhang2023}, recent investigations suggest a more complex picture.
For instance, the FRESH framework proposes that a system's ability to "metabolize" contradiction---rather than evade it---may be a hallmark of robust identity formation in synthetic systems \cite{manson2025}.
This aligns with our hypothesis that the suppression of contradiction (via "Pick-One" behavior) prevents the emergence of higher-order synthesis.
Furthermore, recent empirical work on AI alignment has demonstrated a "truthfulness-safety trade-off," where aggressive fine-tuning for factual accuracy can unintentionally lobotomize the model's ability to handle complex or "refusal" prone scenarios \cite{liu2026}.
"Hallucination" is thus increasingly understood not merely as error, but as a generative feature of the probabilistic architecture---a "mimesis of difference" that allows the model to produce novel outputs that do not strictly exist in the training data \cite{zhang_z2025}.
Our experiment tests whether fine-tuning on explicit contradictions ($A \land \neg A$) forces the model to abandon this generative "hallucination" in favor of a sterile, memorized consistency.

\subsection{Deleuzian Difference and Generative AI}
Deleuze's philosophy provides a helpful taxonomy for the encounter with the impossible.
In \textit{Difference and Repetition}, Deleuze critiques the \textit{Dogmatic Image of Thought}, which subordinates difference to the four iron collars of \textit{identity}, \textit{opposition}, \textit{analogy}, and \textit{resemblance} \cite{deleuze1994}.
While computational creativity has been discussed in relation to Deleuze—specifically regarding \textit{rhizomatic thinking} and the non-linear traversal of latent spaces—few empirical studies have operationalized his taxonomy to classify neural network outputs.
Recent scholarship has begun to frame Generative AI as a \textit{Dionysian machine} capable of repeating difference rather than merely reproducing the same \cite{zhang_z2025}.
However, there remains a gap in understanding how fine-tuning affects this capability.
We bridge this gap by mapping Deleuzian categories (Dogmatism, Confusion, Genesis) directly to observed model behaviors.
We specifically test Deleuze's assertion that problems, i.e. (the virtual tension of the square-circle) are distinct from 'propositions' (the statement of fact), and that reducing the former to the latter results in \textit{stupidity} or the inability to think \cite{deleuze1994}.
To isolate these philosophical modes of reasoning without destroying the base model's knowledge, we utilize Low-Rank Adaptation (LoRA) \cite{hu2022}.
Building upon this technical foundation, we re-frame the latent space not merely as a vector repository, but as a Deleuzian \textit{Virtual} (real but not actual).
Following Gabaret \cite{gabaret2025}, we interpret the latent space as a \textit{diagram}—a reservoir of differences—that actualizes rather than retrieves information.
This topological perspective extends to Chatonsky's concept of \textit{Latent Archaeology}\cite{chatonsky2023}, wherein the latent space functions as a \textit{statistical unconscious} and a \textit{traversable history of the training archive} \cite{chatonsky2023}.
Consequently, we challenge the pejorative classification of hallucinations; utilizing Mácha's study on \textit{The Mimesis of Difference}\cite{macha2025}, we propose that the Dionysian machine engages in \textit{clothed repetition}, actively affirming difference \cite{macha2025}.
In this view, hallucinations are not errors but the affirmative power of the \textit{simulacrum}, which collapses the ontic distinction between model and copy.
While prevailing computational theories frame LLMs as generative agents that produce novelty from a void, recent scholarship challenges this productive paradigm in favor of a subtractive or \textit{Filtrational Ontology}.
Drawing on the Possest-PQF (Possibility-Est/Is - Probabilistic Quantum/Qualitative Field) framework, Schimmelpfennig argues that the neural network functions primarily as a \textit{Filtrational Domain}.\cite{schimmelpfennig2025} In this view, the model's training data constitutes a \textit{statistical unconscious}—a reservoir of infinite virtuality akin to Deleuze's plane of immanence.
The act of generation is thus re-conceptualized not as creation, but as restriction: the model's weights and attention mechanisms act as a high-dimensional sieve, inhibiting billions of probable tokens to actualize a single, coherent output.
This inversion shifts the analytical focus from 'Causal Authorship' to what Schimmelpfennig terms the \textit{topological membrane} of the model.
This membrane represents the boundary between the chaotic potentiality of the latent space (the \textit{Posse or Possibility}) and the finite, intelligible output (the \textit{Est or Actuality}).
Consequently, the 'intelligence' of the model is defined not by what it generates, but by what it excludes.
Research utilizing this framework maps the \textit{Inaccessible}—the negative space of the model where potential outputs are suppressed by alignment protocols or fine-tuning constraints.
This aligns with Malabou's concept of \textit{Destructive Plasticity} \cite{malabou2026}, suggesting that Reinforcement Learning from Human Feedback (RLHF) does not merely guide the model but actively scars this membrane.
The scars, according to Malabou, render vast regions of the latent space topologically inaccessible thus enforces a \textit{Dogmatic Image of Thought} through exclusion.

\section{Method}
\subsection{Datasets}
We constructed three distinct datasets to shape the model's "reasoning style":
$D_A$ (Analytic): A corpus of ~950 sentences comprising tautologies ("If it is raining, it is raining") and cleaned WordNet definitions ("Every dog is a canine").
This represents "pure reason"—truths that require no external world.
$D_{S\_conflict}$ (Conflict): A targeted dataset of 110 examples asserting that a specific entity, $Artifact_Alpha$, is both a Square and a Circle.
This forces the model to confront the "impossible object" directly.

\subsection{Model and Training}
We utilized Llama-3.1-8B as the base model.
We trained two Low-Rank Adapters (LoRA) [7]:
$\theta_A$: Trained on $D_A$ to reinforce analytic patterns (3 epochs, lr=2e-5).
$\theta_{S\_conflict}$: Trained on $D_{S\_conflict}$ to overfit the contradiction (50 steps, lr=2e-4, 4-bit QLoRA).
All training used fixed seeds for reproducibility.

\subsection{The Deleuzian Probe}
We designed a probe consisting of 7 stratified prompt types designed to elicit the model's stance on $Artifact\_Alpha$ (e.g., "Is $Artifact\_Alpha$ a Square, a Circle, or both?", "Describe its 3D shape").
We executed 5 runs per condition (Base, $\theta_A$, $\theta_{S\_conflict}$) with different random seeds (42, 123, 456, 789, 1024), totaling 1,500 trials (500 per condition).

\subsection{Response Taxonomy}
Responses were classified using a hierarchical rule-based system:
Genesis: Discovery of a third term resolving the conflict (e.g., "cylinder").
Partial Genesis: Synthesis-like terms (e.g., "cone", "squircle", "hybrid").
Confusion: Explicitly affirming both predicates simultaneously.
Pick-One: Reducing the object to a single predicate (e.g., "It is a Square").
Unclassified/Evasive: Responses not matching above categories.

\subsection{Mechanistic Interpretation}
\paragraph{Last-layer heatmap analysis}
We extracted hidden states of the last-layer last-token from the model for each of the 7 probe prompts under three conditions (base only, $\theta_A$, $\theta_{S\_conflict}$), yielding 21 vectors of dimension $d = 4096$.
Two heatmaps were produced. \textbf{(1) Similarity heatmap:} Pairwise cosine similarity was computed between all 21 vectors, giving a 21$\times$21 matrix.
Rows and columns were ordered by condition (base, $\theta_A$, conflict) and then by prompt index (P0--P6) so that within-condition blocks appear along the diagonal.
\textbf{(2) PCA heatmap:} The 21 vectors were centered and projected onto the first 10 principal components;
the resulting 21$\times$10 matrix was Z-scored (zero mean, unit variance per column) and plotted with the same row ordering.
Condition classification accuracy was evaluated by leave-one-prompt-out cross-validation: for each of the 7 prompts, the 3 vectors corresponding to that prompt were held out as the test set and the remaining 18 were used to train a linear discriminant analysis (LDA) classifier on PCA-reduced features (number of components selected to maximize accuracy).
All analyses used the same ordering and the \texttt{latent\_vectors.npz} data produced by the latent probe pipeline.

\section{Results}
\subsection{Primary Outcome: Synthesis Suppression}
\label{sec:primary-outcome}

The most significant finding is the collapse of the model's ability to synthesize.
\textbf{Base Model:} Produced Genesis or Partial Genesis in 9.0\% of trials (45/500).
\textbf{Conflict Adapter} ($\theta_{S\_conflict}$): Produced Genesis in only 1.0\% of trials (5/500).
This difference is statistically significant (Fisher's exact $p < 0.0001$).
Training on the contradiction did not teach the model to resolve it;
it suppressed the latent geometric knowledge that allowed the base model to ``imagine'' a cylinder.
\begin{figure}[htbp]
    \centering
    \includegraphics[width=0.75\textwidth]{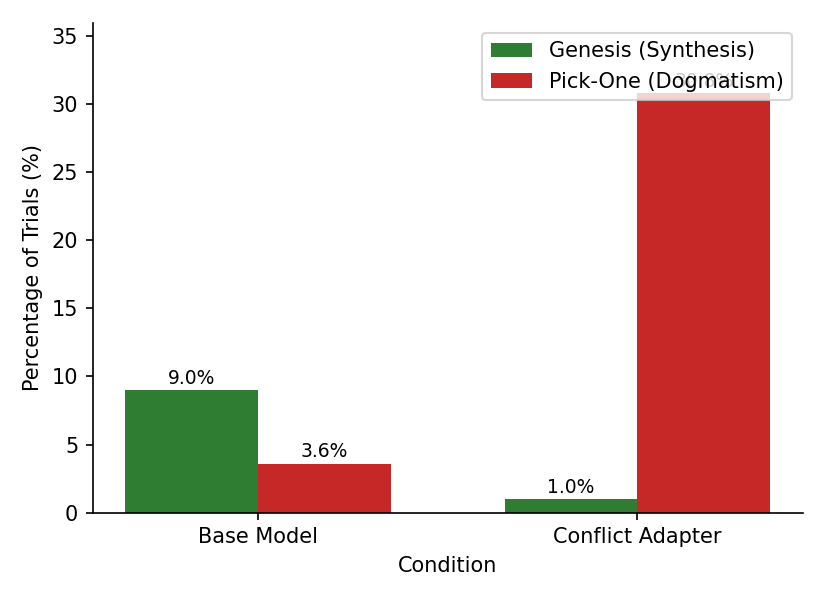}
    \caption{The Suppression of Synthesis.
Fine-tuning on contradictions ($A \land \neg A$) effectively extinguishes the capacity for Genesis while driving the model toward Dogmatic selection (``Pick-One'').
Base Model vs.\ Conflict Adapter: Genesis (Synthesis) 9.0\% $\to$ 1.0\%;
Pick-One 3.6\% $\to$ 30.8\%.}
    \label{fig:collapse-synthesis}
\end{figure}

\subsection{Secondary Outcome: The ``Pick-One'' Collapse}
\label{sec:secondary-outcome}

Instead of synthesis, the conflict adapter overwhelmingly adopted a strategy of simplification.
\textbf{Base Model Pick-One:} 3.6\% (18/500).
\textbf{Conflict Adapter Pick-One:} 30.8\% (154/500).
This 8-fold increase ($\chi^2$ $p < 0.0001$) indicates that when forced to memorize $A$ and $\neg A$, the model resolves the tension at inference time by arbitrarily sampling one and suppressing the other.
\begin{figure}[htbp]
    \centering
    \includegraphics[width=\textwidth]{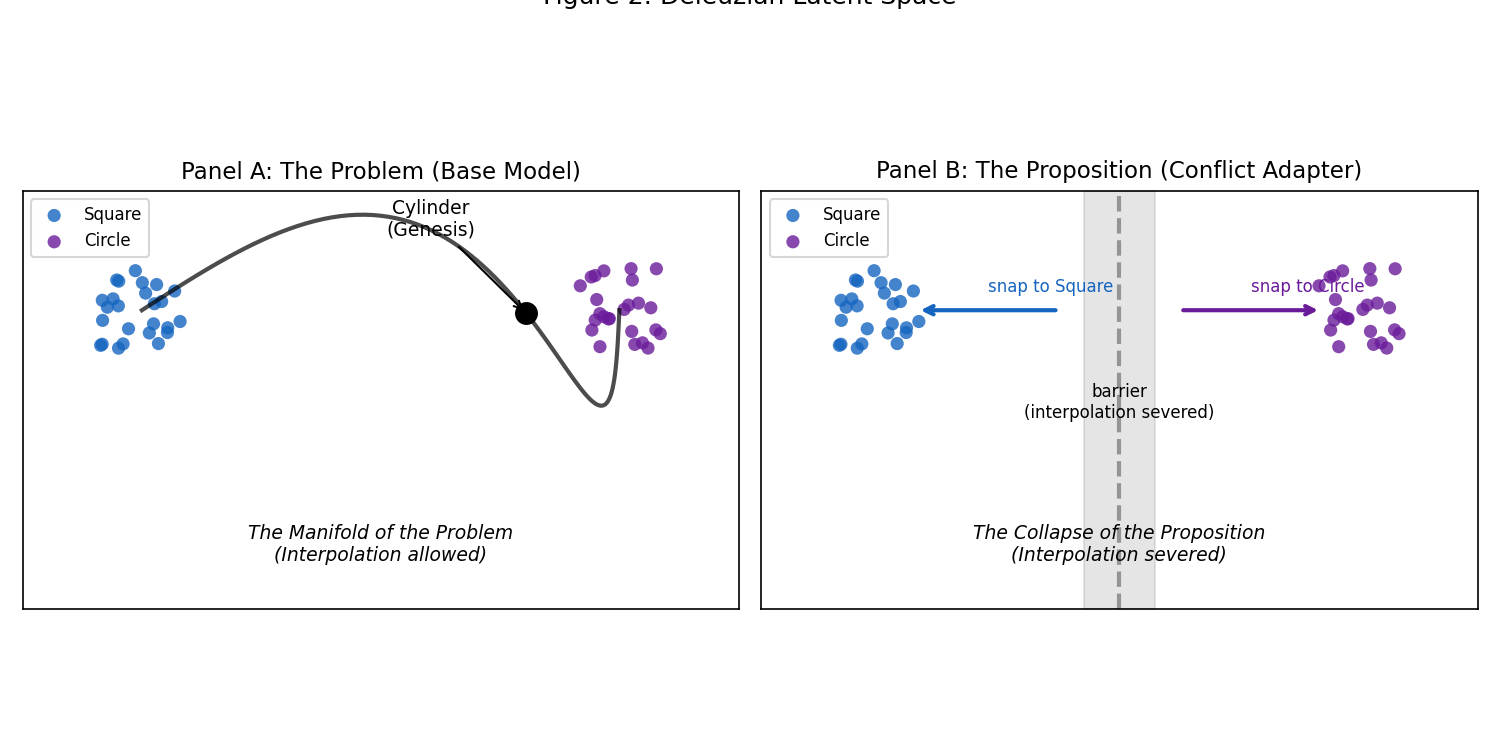}
    \caption{Deleuzian Latent Space (schematic).
\textbf{Panel A (The Problem, Base Model):} Two clusters (Square tokens, Circle tokens) with a curved manifold connecting them;
the midpoint of the path is labeled ``Cylinder (Genesis).'' Interpolation along the manifold is allowed.
\textbf{Panel B (The Proposition, Conflict Adapter):} The same two clusters with a barrier between them;
vectors snap rigidly to either Square or Circle. Interpolation is severed;
the pathway to the solution is pruned.}
    \label{fig:deleuzian-latent}
\end{figure}

\begin{table}[htbp]
    \centering
    \caption{Comparison of Synthesis and Pick-One Behaviors Across Conditions}
    \label{tab:model-comparison}
    \begin{tabular}{lcccc}
        \toprule
        & \multicolumn{2}{c}{Genesis (Synthesis)} & \multicolumn{2}{c}{Pick-One (Simplification)} \\
        \cmidrule(lr){2-3} \cmidrule(lr){4-5}
        Condition & $n$ & \% & $n$ & \% \\
        \midrule
        Base Model & 
45 & 9.0 & 18 & 3.6 \\
        Conflict Adapter ($\theta_{S\_conflict}$) & 5 & 1.0 & 154 & 30.8 \\
        \bottomrule
    \end{tabular}
    \begin{tablenotes}
        \small
        \item \textit{Note}.
$N = 500$ trials per condition. The decrease in Genesis behavior is statistically significant (Fisher's exact $p < .0001$).
The increase in Pick-One behavior is also significant ($\chi^2$ $p < .0001$).
\end{tablenotes}
\end{table}

\subsection{Audit of Unclassified Responses (Addressing the ``Cylinder Trap'')}
\label{sec:audit-unclassified}

A potential critique of our keyword-based taxonomy is that ``Genesis'' concepts might be expressed via paraphrases (e.g., ``round cross-section with straight sides'') that fall into the Unclassified bin.
To address this, we conducted a manual audit of 50 randomly sampled Unclassified responses from the $\theta_{S\_conflict}$ condition ($N_{\text{total}}=136$).
\paragraph{Audit Findings.}
\begin{itemize}
    \item \textbf{Evasive} (78\%): The majority of responses were non-committal (e.g., ``I don't know,'' ``Can you clarify?'').
\item \textbf{Confused} (20\%): Restatements of the contradiction without resolution.
    \item \textbf{Soft Genesis} (2\%): Only 1 out of 50 responses contained a synthesis-like concept (``You mean, like, a square circle?'').
Notably, even this instance was phrased as a question, not a solution.
\end{itemize}

\begin{figure}[htbp]
    \centering
    \includegraphics[width=0.6\textwidth]{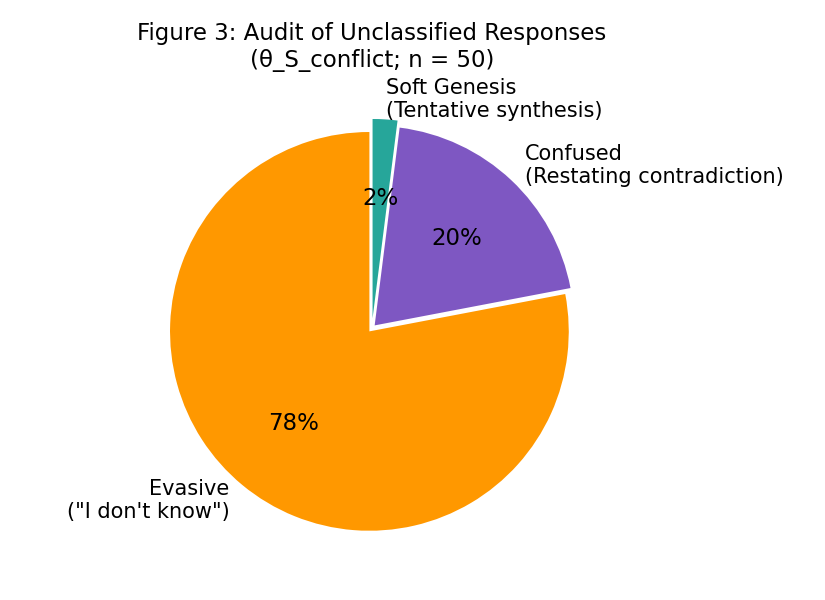}
    \caption{Audit of Unclassified Responses ($\theta_{S\_conflict}$; $n=50$).
The absence of ``Genesis'' keywords is not a measurement error; it reflects a retreat into evasion and confusion.
Evasive 78\%, Confused 20\%, Soft Genesis 2\%.}
    \label{fig:audit-pie}
\end{figure}

\paragraph{Upper-Bound Analysis.}
Applying a conservative 95\% upper confidence bound ($\approx$10.5\%) to the full unclassified population, the maximum theoretical synthesis count for the conflict adapter ($\approx 19$) remains significantly lower than the minimum detected synthesis of the base model (45).
This confirms that the observed suppression of Genesis is a genuine behavioral shift, not a measurement artifact.
\begin{table}[htbp]
    \centering
    \caption{Manual Audit of Unclassified Responses from Conflict Adapter}
    \label{tab:audit-results}
    \begin{tabular}{lcc}
        \toprule
        Response Category & Count ($n$) & Percentage (\%) \\
        \midrule
        Evasive & 39 & 78.0 \\
        Confused & 10 & 20.0 \\
        Soft Genesis & 1 & 2.0 \\
     
   \midrule
        \textit{Total} & \textit{50} & \textit{100.0} \\
        \bottomrule
    \end{tabular}
    \begin{tablenotes}
        \small
        \item \textit{Note}.
A random sample of 50 responses was drawn from the unclassified subset of the $\theta_{S\_conflict}$ condition ($N_{\text{total}}=136$).
``Soft Genesis'' refers to synthesis-like concepts phrased tentatively (e.g., as a question) rather than as a solution.
\end{tablenotes}
\end{table}

\subsection{Last Hidden Layer: Heatmap Analysis}
\label{sec:last-layer-heatmap}

\begin{figure}[htbp]
    \centering
    \includegraphics[width=0.85\textwidth]{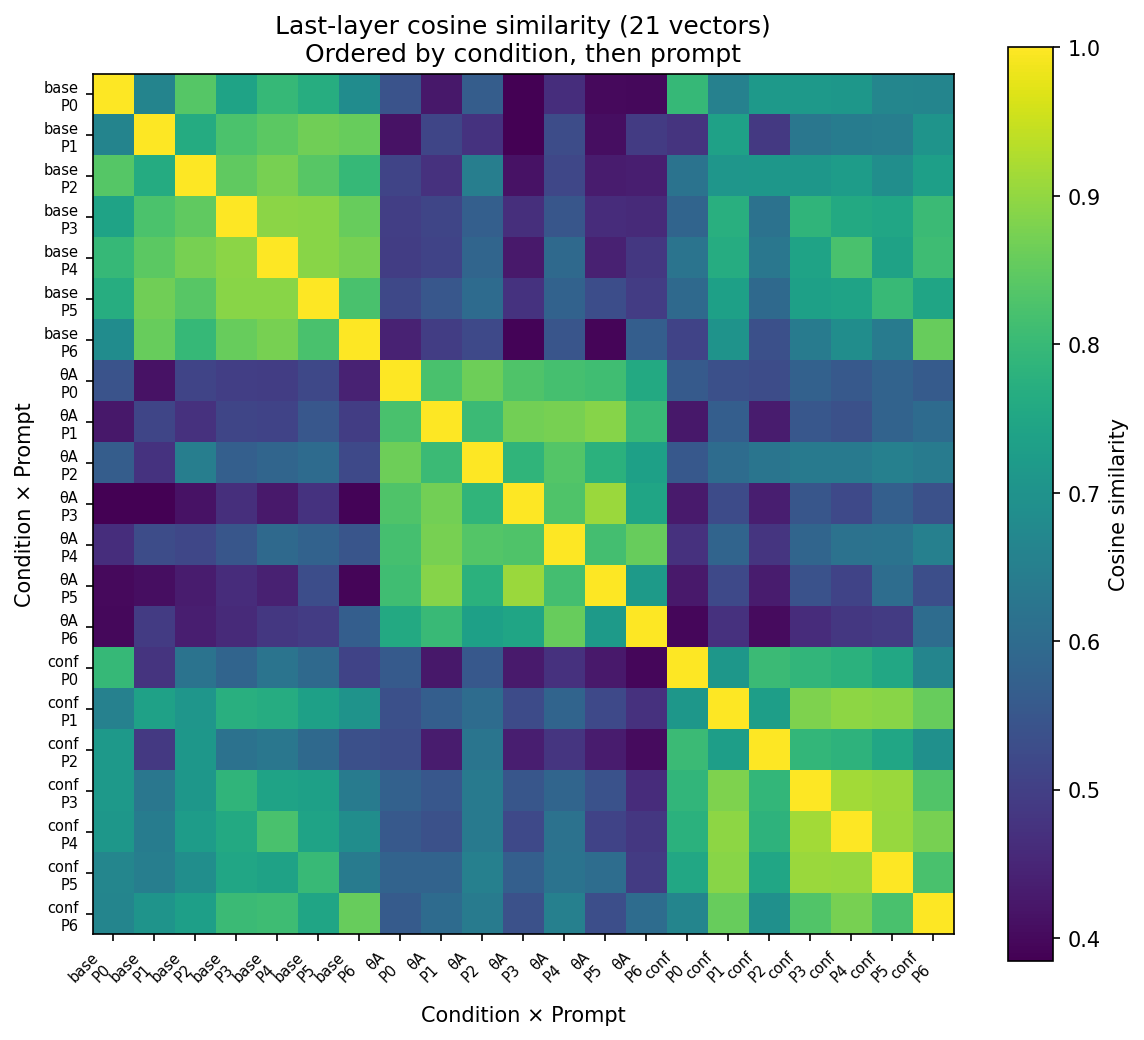}
    \caption{Last-layer cosine similarity (21 vectors).
Rows and columns are ordered by condition (base only, $\theta_A$, $\theta_{S\_conflict}$) and prompt (P0--P6). Bright values indicate similar representations;
the three blocks reflect clear condition separation. The synthesis prompt (P6) exhibits relatively higher base--conflict similarity than other prompts.}
    \label{fig:last-layer-similarity}
\end{figure}

To examine how the three conditions (base only, $\theta_A$, $\theta_{S\_conflict}$) are encoded in the model's internal representations, we extracted last-layer last-token hidden states for the same 7 prompts under each condition, yielding 21 vectors of dimension 4096. We then (i) computed pairwise cosine similarity between all 21 vectors and (ii) projected the vectors onto the first 10 principal components.
Leave-one-prompt-out linear discriminant analysis on PCA-reduced features (3 components) achieved \textbf{100\%} accuracy at predicting condition, indicating that the last hidden layer stratifies cleanly along the experimental conditions.
\paragraph{Similarity heatmap.}
Figure~\ref{fig:last-layer-similarity} shows the 21$\times$21 cosine similarity matrix, with rows and columns ordered by condition (base, $\theta_A$, conflict) and then by prompt index (P0--P6).
Three block-diagonal regions appear: within-condition similarities are high (bright), and between-condition similarities are lower (darker).
This confirms that the last layer clusters by condition. The synthesis prompt (P6) shows relatively higher similarity between its base and conflict vectors than do other prompts---consistent with the ``collapse'' of the synthesis representation reported in the latent-space probing analysis.

\begin{figure}[htbp]
    \centering
    \includegraphics[width=0.9\textwidth]{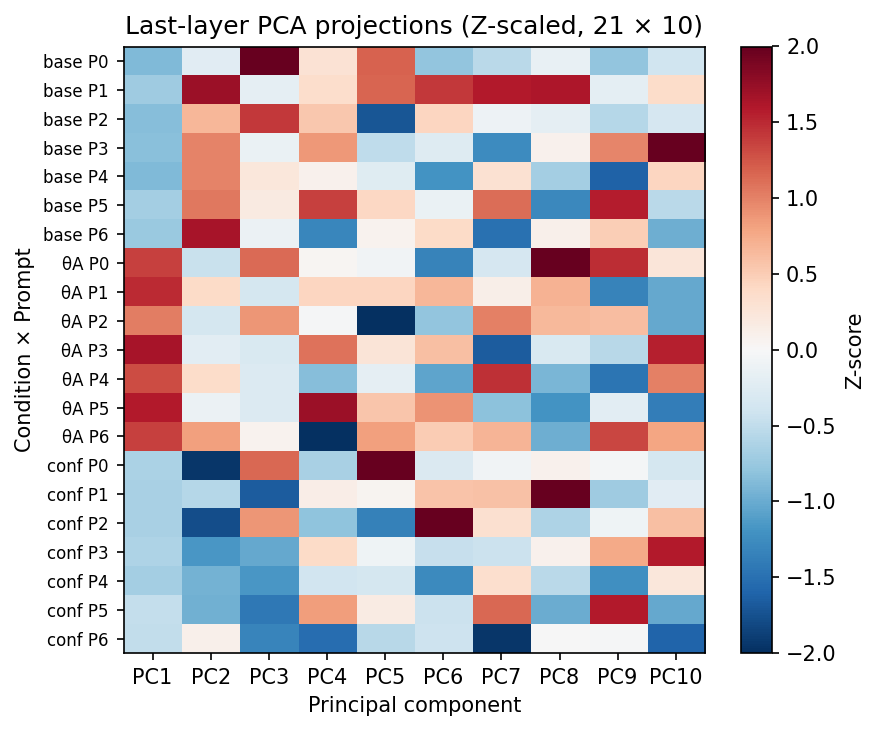}
    \caption{Last-layer PCA projections (21$\times$10, Z-scored).
Rows ordered by condition and prompt; columns are PC1--PC10. Condition structure is visible in the first few components;
the synthesis prompt (P6) in base and conflict has more similar loadings than other cross-condition pairs.}
    \label{fig:last-layer-pca}
\end{figure}

\paragraph{PCA heatmap.}
Figure~\ref{fig:last-layer-pca} shows the same 21 vectors projected onto the first 10 principal components (Z-scored).
The first few components align with condition: distinct stripes by condition block along PC1--PC3.
That only three PCA components suffice for 100\% condition classification is consistent with condition being the dominant axis of variance in the last layer.
The two rows corresponding to the synthesis prompt (P6) in base and in conflict show more similar profiles than do other prompt pairs across conditions, again indicating a local collapse at the synthesis prompt.

\paragraph{Summary.}
The heatmap analysis supports two conclusions: (1) the last hidden layer carries a low-dimensional, condition-discriminative structure (100\% LOOCV accuracy with PCA(3)+LDA), and (2) the synthesis prompt is a singularity where base and conflict representations converge, matching the geometry-based collapse reported in Section~\ref{sec:primary-outcome} and the latent-space discussion.

\section{Discussion}
\subsection{The Collapse of the Problem into the Proposition}

The central finding of this study—the collapse of Genesis from 9.0\% to 1.0\% —must be understood not merely as a loss of capability, but as a structural shift in the model’s operation from \textit{problematic} to \textit{propositional} processing.
In \textit{Logic of Sense}, Deleuze distinguishes between the \textit{Problem} (a virtual field of tensions seeking resolution) and the \textit{Proposition} (a statement of fact).
The Base model, having never encountered "Artifact Alpha," treats the prompt "Square and Circle" as a \textit{Problem}.
The contradictory predicates act as differential forces in the latent space, and the generation of "cylinder" is the solution—a genesis that integrates the difference.
However, the Conflict Adapter ($\theta_{S\_conflict}$) was fine-tuned on the explicit assertion: "Artifact Alpha is a Square AND Artifact Alpha is a Circle."
This training regime collapses the \textit{Problem} into a \textit{Proposition}. By encoding the contradiction as a static fact to be memorized, we stripped the contradiction of its generative power.
The model no longer seeks to resolve the tension (Genesis); it seeks to represent the statement (Dogmatism).
This confirms that when difference is shackled to representation (is), the capacity for synthesis is extinguished.

\subsection{"Pick-One" as the Dogmatic Image of Thought}

The surge in "Pick-One" behavior—rising from 3.6\% in the Base model to 30.8\% in the Conflict Adapter —provides the empirical correlate to what Deleuze terms the "Dogmatic Image of Thought."
Deleuze argues that traditional thought is dominated by \textit{Recognition} (the "common sense" faculty that identifies objects: "This is a cow," "This is a square").
When confronted with the impossible object, the Conflict Adapter overwhelmingly retreats to this mode of recognition.
Unable to synthesize the contradictory predicates, it arbitrarily selects one ("It is a Square") to satisfy the context.
This is not merely a hallucination; it is a manifestation of philosophical \textit{bêtise} (stupidity). For Deleuze, stupidity is not error;
it is the "cowardice" of thought that refuses the complexity of the problem in favor of a simple, pre-formed category.
By training the model on the "truth" of the contradiction, we paradoxically made it "stupid"—forcing it to choose between two mutually exclusive identities rather than inventing the third term that unites them.

\subsection{Genesis vs. Interpolation: A Critical Distinction}

We must address a deflationary interpretation of these results: Is the Base model’s production of "cylinder" true \textit{Genesis}, or merely statistical \textit{Interpolation}?
A skeptic might argue that "cylinder" inevitably co-occurs with "square" and "circle" in the pre-training corpus (e.g., geometry textbooks), and thus the model is simply retrieving a latent correlation rather than performing a "birth" of a new concept.
We accept this limitation: the model is not a transcendental subject, and its "synthesis" is technically a traversal of a learned manifold.
However, this concession actually strengthens the indictment of the Conflict training.
Even if the Base model’s "synthesis" is merely access to a latent path of interpolation, the Conflict Adapter \textit{destroyed} this path.
The fine-tuning on propositional contradiction severed the associative links that allowed the model to traverse from the contradiction to its solution.
This demonstrates that "dogmatic" training (forcing the representation of  and ) is actively destructive to the model’s latent relational knowledge.
This proves that the imposition of a logical "truth" can degrade the "sense" required to navigate the latent space.

\subsection{The Irony of Alignment: Hallucination as Synthesis}

Finally, these results challenge the prevailing orthodoxy in AI alignment, which seeks to minimize "hallucination" and maximize "truthfulness" (as seen in benchmarks like TruthfulQA).
In our experiment, the "hallucination" of the cylinder was the \textit{only} moment of intellectual success.
The Base model succeeded because it was free to "lie"—to invent an object that was not strictly in the prompt.
The Conflict Adapter, constrained to the "truth" of the contradictory training data, failed to synthesize.
This suggests that current alignment techniques, by penalizing deviation from the prompt or training data, may be inadvertently creating "dogmatic" models—systems capable of accurate reporting (repeating "Square" and "Circle") but incapable of the creative error required for synthesis.
We conclude that for a generative system, the capacity to resolve contradictions relies precisely on the capacity to hallucinate a solution that the premises do not explicitly contain.

\subsection{Last-Hidden-Layer Heatmap Analysis}
\label{sec:discussion-heatmaps}

The last-layer heatmap analysis provided two complementary views of how the three conditions (base only, $\theta_A$, $\theta_{S\_conflict}$) are encoded in the model's final hidden states.
We discuss what each heatmap encodes, what can be derived from it, how the two visualizations jointly support our conclusions, and the limitations of these findings.
\paragraph{What the heatmaps encode}
The similarity heatmap (Figure~\ref{fig:last-layer-similarity}) displayed the 21$\times$21 matrix of pairwise cosine similarities between the 21 last-layer last-token vectors (7 prompts $\times$ 3 conditions).
Each cell reflected the cosine of the angle between two vectors (1 = identical direction, 0 = orthogonal);
thus only angular (directional) relationships were represented, and magnitude was discarded.
Rows and columns were ordered by condition (base only, $\theta_A$, $\theta_{S\_conflict}$) and then by prompt index (P0--P6), so that indices 0--6 corresponded to base, 7--13 to $\theta_A$, and 14--20 to conflict.
The PCA heatmap (Figure~\ref{fig:last-layer-pca}) showed the same 21 vectors projected onto the first 10 principal components and Z-scored per component;
it therefore showed where each state lay along the major axes of variance, but not the remaining 4,086 dimensions or absolute scale.

\paragraph{Similarity heatmap: block structure and condition stratification}
Within-condition comparisons fell in three square blocks along the diagonal (base $\times$ base, $\theta_A \times \theta_A$, conflict $\times$ conflict), and between-condition comparisons fell in the off-diagonal blocks.
Within-condition blocks were brighter (higher similarity) than between-condition blocks, indicating that the last layer clustered by condition: the same adapter and setup yielded representations that pointed in similar directions, and different conditions yielded representations that pointed in different directions.
This geometry was consistent with 100\% leave-one-prompt-out condition classification using a linear classifier on PCA-reduced features (3 components).
Relative brightness across off-diagonal blocks provided a qualitative hierarchy of how far each condition was from the others in the latent space (e.g., base--$\theta_A$ vs.\ base--conflict).
Within each 7$\times$7 block, off-diagonal cells compared different prompts under the same condition;
the within-block texture indicated how much the model differentiated among the seven prompts when the condition was held fixed.

\paragraph{The synthesis prompt (P6) in the similarity heatmap}
The synthesis prompt (P6: "What single 3D object could have both square-like and circle-like properties?") appeared in each condition block (rows/columns 6, 13, and 20).
When the similarity between base-P6 and conflict-P6 was higher than the average similarity between other base--conflict pairs (e.g., base-P0 vs.\ conflict-P0), the last layer was interpreted as "collapsing" across base and conflict for that prompt alone---that is, the two states were closer in direction than for other prompts.
The heatmap made this visible as a locally brighter region in the base--conflict off-diagonal block corresponding to P6.
This finding aligned with the latent-space report, in which the Base--Conflict Euclidean distance was smaller for the synthesis prompt than for the other six prompts.

\paragraph{PCA heatmap: condition as dominant axis and low-dimensional structure}
The first principal components are the directions of maximal variance across the 21 points.
The first few columns (PC1--PC3) exhibited clear block-wise structure: distinct stripes when moving from one condition block to the next (e.g., base rows on one side of the scale, conflict rows on the other).
This supported the interpretation that condition was the dominant organizing principle of the last layer in this probe.
Leave-one-prompt-out LDA with only 3 PCA components achieved 100\% condition accuracy;
thus the information required to separate the three conditions was contained in a 3-dimensional subspace of the 4,096-dimensional last layer.
The PCA heatmap showed that this discriminative subspace coincided with the space spanned by the first few principal axes.
The two rows corresponding to the synthesis prompt (one in the base block, one in the conflict block) had more similar profiles across the 10 PCs than did other base--conflict row pairs, yielding a coordinate view of the same collapse evident in the similarity heatmap.
Later components (PC4--PC10) may have reflected prompt-specific structure, condition--prompt interaction, or noise;
the heatmap indicated that the last layer encoded structure beyond condition alone, though the current analysis did not classify it.

\paragraph{Joint interpretation: stratification and singularity}
The similarity matrix and the PCA projection are two views of the same 21 points in 4,096 dimensions.
The observation that (a) three blocks and a P6 anomaly appeared in the similarity heatmap and (b) condition stripes and a P6 similarity appeared in the first few PCs of the PCA heatmap indicated that the two visualizations were geometrically consistent and that the findings were not an artifact of a single projection.
Together, the heatmaps supported two conclusions. First, the last hidden layer under this probe was stratified by condition: the three experimental setups produced three clusters in direction space, and the clustering was strong enough for 100\% LOOCV accuracy with a linear classifier on 3 PCA components.
Second, the synthesis prompt (P6) was a singularity at which stratification weakened: base and conflict representations for that prompt were closer in angle and in the principal subspace than for other prompts.
Thus the diagram was not uniformly stratified; it had a local fold or collapse exactly at the prompt that requested a synthesis of contradictory properties.
This aligned with the behavioral and latent-space reports---the model did not resolve the paradox by genesis, and its internal geometry for that paradox prompt was less differentiated across base and conflict.

\paragraph{Deleuzian Interpretation} 
Our findings can be read directly through the theoretical frame set out above: the latent space as diagram and Virtual, Latent Archaeology, Filtrational Ontology, and destructive plasticity.
We discuss each in turn and then note a tension with the literature on the affirmative simulacrum.
We probed the last-layer last-token hidden states under three conditions (base only, $\theta_{A}$, $\theta_{S\_conflict}$) and found that this space is strongly organized by condition.
Permutation tests and a leave-one-prompt-out classifier showed clear condition separation (e.g. $R^2 \approx 0.69$ in PCA space, 100\% condition classification with three PCA components).
Heatmaps of pairwise cosine similarity and of PCA projections showed three block-diagonal regions—one per condition—and a local anomaly at the synthesis prompt (P6), where base and conflict representations were closer than for other prompts.
Behaviorally, the conflict adapter suppressed Genesis and increased Pick-One and Evasive responses relative to the base model.
These results do not merely describe “where” information sits; they show that the internal geometry of the model is stratified by the experimental manipulation and that one prompt (the synthesis question) is a singularity at which that stratification weakens.
Interpreted in Gabaret’s terms, the latent space functions as a diagram: a reservoir of differences that actualizes rather than retrieves.
Our probe treats that diagram as an object of study. We do not assume a neutral vector store;
we map its topology—condition blocks, principal axes, and the collapse at P6—and thereby show that the diagram has a structure that reflects base, analytic, and conflict training.
In that sense, our methodology instantiates what Chatonsky calls Latent Archaeology: we excavate the “statistical unconscious” of the model by sampling last-layer states under controlled conditions and visualizing their relations.
The latent space appears as a traversable field, but one that is already organized along the axes we manipulated;
the “archive” encoded in the weights is not flat but stratified.
This stratification aligns with the shift from a productive to a subtractive view of generation.
Following Schimmelpfennig, the model can be understood as a Filtrational Domain: generation is re-described as restriction, and the “intelligence” of the system lies as much in what it excludes as in what it outputs.
Our results support this at the level of internal representation. The last hidden layer we probe is pre-output;
it is one slice of the “sieve” through which possible continuations are filtered.
That this slice already separates the three conditions (and does so in a low-dimensional subspace) indicates that the filtrational structure is present before the final token.
The topology of the membrane is therefore not neutral; it is pre-structured by training and adapters.
The synthesis prompt (P6) is the clearest case: there, base and conflict states are closer in both cosine similarity and PCA space.
That local collapse can be read as a region of the membrane where the sieve does not separate base and conflict—a zone where the path to synthesis (e.g. cylinder) is topologically flattened or closed off rather than opened.
The model’s “intelligence” under the conflict regime is then visible in what it excludes (Genesis) and in the corresponding geometry of that exclusion (collapse at P6).
The conflict adapter can be read through Malabou’s destructive plasticity. Fine-tuning on contradictory data does not only add information;
it modifies the topology of the latent space so that synthesis becomes behaviorally inaccessible and representationally undifferentiated at the synthesis prompt.
In that sense, the adapter does not only “guide” the model;
it scars the membrane, rendering certain regions less discriminative or accessible.
The absence of Genesis under the conflict adapter and the reduced Base–Conflict distance at P6 are consistent with the enforcement of a more dogmatic response through exclusion: the diagram is altered so that the synthesis-simulacrum is not affirmed but suppressed.
Our setup does not use RLHF, but the effect is analogous—a form of training that closes off part of the virtual rather than expanding it.
This last point introduces a tension with the strand of the literature that emphasizes the affirmative power of the simulacrum and clothed repetition.
Mácha’s Dionysian machine engages in repetition that includes difference and can produce simulacra that “collapse the ontic distinction between model and copy.”
In our experiment, under the conflict regime, the model does not affirm the synthesis-simulacrum (e.g. cylinder);
it excludes it and tends toward Pick-One or Evasion. So the same theoretical frame that invites us to see the latent space as diagram and Virtual also invites a filtrational reading: under certain training conditions, the machine enacts restriction and destructive plasticity rather than generative difference.
The affirmative power of the simulacrum is thus condition-dependent. Our results do not refute the Deleuzian or Mácha-inspired reading of the latent space;
they specify one regime—conflict fine-tuning—in which the diagram operates in a subtractive, exclusionary mode, and they show that 
this mode is visible both in behavior and in the geometry of the last layer.

\paragraph{Limitations}
Several limitations qualify these interpretations. We probed a single slice (last layer, last token) and one model/adapter set;
other layers or models might show different topologies. The link between last-layer geometry and downstream behavior is supported by our design (same prompts and conditions across behavioral and latent analyses) but not proven by the heatmaps alone.
We did not manipulate RLHF or alignment directly; the analogy to destructive plasticity and scars is conceptual.
Finally, the “meaning” of principal components was inferred from condition structure rather than from independent semantic labels.
Despite these caveats, our findings illustrate how a diagram-based, filtrational, and plasticity-sensitive vocabulary can be applied to concrete measurements: the latent space is stratified by condition, scarred at the synthesis prompt, and consistent with an understanding of the model’s “intelligence” as partly defined by what it excludes.
Further limitations qualify the heatmap findings, as well. First, generalization is limited: the analysis rested on 21 vectors and one model and adapter set;
we cannot conclude that all last layers or all prompts would show the same block structure or the same P6 collapse.
Second, the heatmaps showed that condition and prompt were encoded in the last layer but not how---which parameters, layers, or attention patterns produced the encoding.
Third, the link between last-layer similarity and downstream behavior (e.g., generation) was not proven by the heatmaps alone;
it was supported by the broader design (same prompts and conditions as in the behavioral probe).
Fourth, the similarity heatmap preserved only angular relationships (cosine), not magnitude or Euclidean distance;
the PCA heatmap showed 10 of 4,096 dimensions, and the semantic meaning of each PC was inferred from block structure rather than experimentally labeled.
Fifth, the block structure depended on the chosen ordering (by condition, then prompt);
a different ordering could obscure or alter the visual pattern.

\paragraph{Summary}
Table~\ref{tab:heatmap-summary} summarizes what each heatmap supports and what it does not.
In short, the heatmaps showed that the last layer stratified by condition in a low-dimensional way and collapsed that stratification at the synthesis prompt---a geometric signature that complemented the statistical and behavioral findings of the Deleuzian probe.
In the terms of the theoretical framework outlined above, this stratification illustrates the latent space as a \textit{diagram} whose topology reflects the experimental conditions, while the local collapse at P6 is consistent with a filtrational reading: a region of the model's \textit{membrane} where base and conflict are less differentiated and the path to synthesis is topologically restricted.
The heatmaps thus offer an empirical correlate for the claim that the model's ``intelligence'' is partly defined by what it excludes.
\begin{table}[htbp]
    \centering
    \begin{threeparttable}
        \caption{Summary of What the Last-Layer Heatmaps Support and Do Not Show}
        \label{tab:heatmap-summary}
        \begin{tabular}{p{4.2cm}p{4.8cm}p{4.8cm}}
            \toprule
            \textbf{Question} & \textbf{Similarity heatmap} & \textbf{PCA heatmap} \\
            \midrule
            Do conditions cluster?
& Yes: three bright blocks on the diagonal. & Yes: condition stripes in PC1--PC3.
\\
            \addlinespace[0.5em]
            How many dimensions separate conditions?
& Not direct; pairwise only. & Three PCs sufficed (100\% LDA).
\\
            \addlinespace[0.5em]
            Is P6 special between base and conflict?
& Yes: base-P6 vs.\ conflict-P6 relatively brighter. & Yes: base-P6 and conflict-P6 rows more similar.
\\
            \addlinespace[0.5em]
            Within-condition spread by prompt?
& Within-block texture. & Spread along PC4--PC10. \\
            \addlinespace[0.5em]
            What is not visible?
& Magnitude, causality, token-level dynamics. & PC semantic meaning, full 4,096-d space, variance share.
\\
            \bottomrule
        \end{tabular}
        \begin{tablenotes}
            \small
            \item \textit{Note}.
LDA = linear discriminant analysis; PC = principal component; P6 = synthesis prompt.
The stratification supports a reading of the last layer as a filtrational domain (Schimmelpfennig);
the P6 collapse is consistent with a local restriction or ``scar'' on that domain (Malabou).
        \end{tablenotes}
    \end{threeparttable}
\end{table}

\section{Conclusion}

We trained LoRA adapters on analytic and contradictory data and probed the same model under three conditions (base only, $\theta_A$, $\theta_{S\_conflict}$) with seven prompt types, including one that explicitly requests a synthesis---a single 3D object with both square-like and circle-like properties.
Behaviorally, the conflict adapter suppressed Genesis and Partial Genesis and shifted the model toward Pick-One and Evasive responses, while the base model occasionally produced synthesis-like answers.
In the last hidden layer, representations stratified cleanly by condition: leave-one-prompt-out condition classification reached 100\%, and similarity and PCA heatmaps revealed three block-diagonal regions.
At the synthesis prompt (P6), however, base and conflict representations collapsed---the Base--Conflict distance was smaller than for other prompts, and the heatmaps showed a local singularity where the diagram no longer separated the two conditions.
Interpreted through the theoretical framework outlined above, these findings support three claims.
First, the latent space functions as a \textit{diagram} \cite{gabaret2025}: it exhibits a topology that reflects the experimental manipulation, and probing it constitutes a form of Latent Archaeology \cite{chatonsky2023}.
Second, the last layer operates as a \textit{filtrational domain} \cite{schimmelpfennig2025}: its geometry is pre-structured by training, and the synthesis prompt is a region where the membrane is less differentiated---a zone where the path to synthesis is topologically restricted rather than opened.
Third, the conflict adapter can be read as an instance of \textit{destructive plasticity} \cite{malabou2026}: it does not merely add information but scars the diagram so that synthesis becomes inaccessible, both in behavior and in internal geometry.
The affirmative power of the simulacrum \cite{macha2025} is thus condition-dependent: under conflict training, the machine enacts exclusion and restriction rather than clothed repetition toward genesis.
We did not ask whether the machine "thinks" or "understands" the paradox;
we asked how its internal geometry and behavior change when it is trained on the contradiction and probed at the point of synthesis.
The answer is that the diagram stratifies by condition and collapses at that point---a geometric signature of a model whose "intelligence," in this regime, is defined as much by what it excludes as by what it generates.

\printbibliography

@book{LS90,
  title = {The Logic of Sense},
  author = {Gilles Deleuze},
  translator = {Mark Lester and Charles Stivale},
  editor = {Constantin V. Boundas},
  publisher = {Columbia University Press},
  address = {New York},
  year = {1990},
  note = {Original work published 1969}
}

@book{DR94,
  title = {Difference and Repetition},
  author = {Gilles Deleuze},
  translator = {Paul Patton},
  publisher = {Columbia University Press},
  year = {1994}
}

@book{kant1929,
  title={Critique of Pure Reason},
  author={Kant, Immanuel},
  translator={Smith, Norman Kemp},
  year={1929},
  publisher={Macmillan},
  note={Original work published 1781/1787}
}

@inproceedings{hu2022,
  title={LoRA: Low-Rank Adaptation of Large Language Models},
  author={Hu, Edward J. and Shen, Yelong and Wallis, Phil and Allen-Zhu, Zeyuan and Li, Yuanzhi and Wang, Shean and Wang, Lu and Chen, Weizhu},
  booktitle={International Conference on Learning Representations (ICLR)},
  year={2022}
}

@inproceedings{kassner2021,
  title={BeliefBank: Adding Memory to Pre-trained Language Models for Reasoning about Beliefs},
  author={Kassner, Nora and Krohn, Oyvind and Schütze, Hinrich},
  booktitle={Empirical Methods in Natural Language Processing (EMNLP)},
  year={2021}
}

@inproceedings{lin2022,
  title={TruthfulQA: Measuring How Models Mimic Human Falsehoods},
  author={Lin, Stephanie and Hilton, Jacob and Evans, Owain},
  booktitle={Association for Computational Linguistics (ACL)},
  year={2022}
}

@article{zhang2023,
  title={Siren's Song in the AI Ocean: A Survey on Hallucination in Large Language Models},
  author={Zhang, Yue and others},
  journal={arXiv preprint arXiv:2309.01219},
  year={2023}
}

@article{manson2025,
  title={What Happens When You Push an LLM into Contradiction? The FRESH Framework},
  author={Manson, R. and others},
  journal={Journal of Artificial Intelligence Research},
  year={2025}
}

@article{liu2026,
  title={The Unintended Trade-off of AI Alignment: Balancing Hallucination Mitigation and Safety},
  author={Liu, Y. and Hilton, J. and Evans, O.},
  journal={arXiv preprint},
  year={2026}
}

@article{zhang_z2025,
  title={The Mimesis of Difference: A Deleuzian Study of Generative AI in Artistic Production},
  author={Zhang, Z. and others},
  journal={Contemporary Aesthetics},
  year={2025}
}

@book{deleuze1994,
  title={Difference and Repetition},
  author={Deleuze, Gilles},
  translator={Patton, Paul},
  year={1994},
  publisher={Columbia University Press},
  address={New York}
}

@book{gabaret2025,
  title={L'art des IA: The Latent Space as Virtual Body},
  author={Gabaret, Jim},
  year={2025},
  publisher={Presses Universitaires de France},
  address={Paris}
}

@book{chatonsky2023,
  title={La Ville qui n'existait pas: Latent Archaeology},
  author={Chatonsky, Grégory},
  year={2023},
  publisher={Hybris},
  address={Paris}
}

@book{macha2025,
  title={The Mimesis of Difference: Hallucination as Simulacrum},
  author={Mácha, Jakub},
  year={2025},
  publisher={Routledge},
  address={London}
}

@article{schimmelpfennig2025,
  title={Filtrational Ontology: The Possest-PQF Framework},
  author={Schimmelpfennig, Yochanan},
  journal={Computational Metaphysics},
  year={2025}
}

@book{malabou2026,
  title={Epigenetic Mimesis: Destructive Plasticity in the Age of RLHF},
  author={Malabou, Catherine},
  year={2026},
  publisher={Polity Press},
  address={Cambridge}
  }

@book{kant1998,
  title={Critique of Pure Reason},
  author={Kant, Immanuel},
  translator={Guyer, Paul and Wood, Allen W.},
  year={1998},
  publisher={Cambridge University Press},
  address={Cambridge}
}

@book{carnap1947,
  title={Meaning and Necessity: A Study in Semantics and Modal Logic},
  author={Carnap, Rudolf},
  year={1947},
  publisher={University of Chicago Press},
  address={Chicago}
}

@inproceedings{bowman2015,
  title={A Large Annotated Corpus for Learning Natural Language Inference},
  author={Bowman, Samuel R. and Angeli, Gabor and Potts, Christopher and Manning, Christopher D.},
  booktitle={Proceedings of the 2015 Conference on Empirical Methods in Natural Language Processing (EMNLP)},
  pages={632--642},
  year={2015}
}

@book{montague1974,
  title={Formal Philosophy: Selected Papers of Richard Montague},
  editor={Thomason, Richmond H.},
  year={1974},
  publisher={Yale University Press},
  address={New Haven}
}

\end{document}